# An Integrated Transfer Learning and Multitask Learning Approach for Pharmacokinetic Parameter Prediction

Zhuyifan Ye[1], Yilong Yang[1,2], Xiaoshan Li[2], Dongsheng Cao[3], Defang Ouyang[1*]

[1]State Key Laboratory of Quality Research in Chinese Medicine, Institute of Chinese Medical Sciences (ICMS), University of Macau, Macau, China

[2]Department of Computer and Information Science, Faculty of Science and Technology, University of Macau, Macau, China

[3]Xiangya School of Pharmaceutical Sciences, Central South University, No. 172, Tongzipo Road, Yuelu District, Changsha, People's Republic of China

Note: Zhuyifan Ye and Yilong Yang made equal contribution to the manuscript.

Corresponding author: Defang Ouyang, email: defangouyang@umac.mo; telephone: 853-88224514.

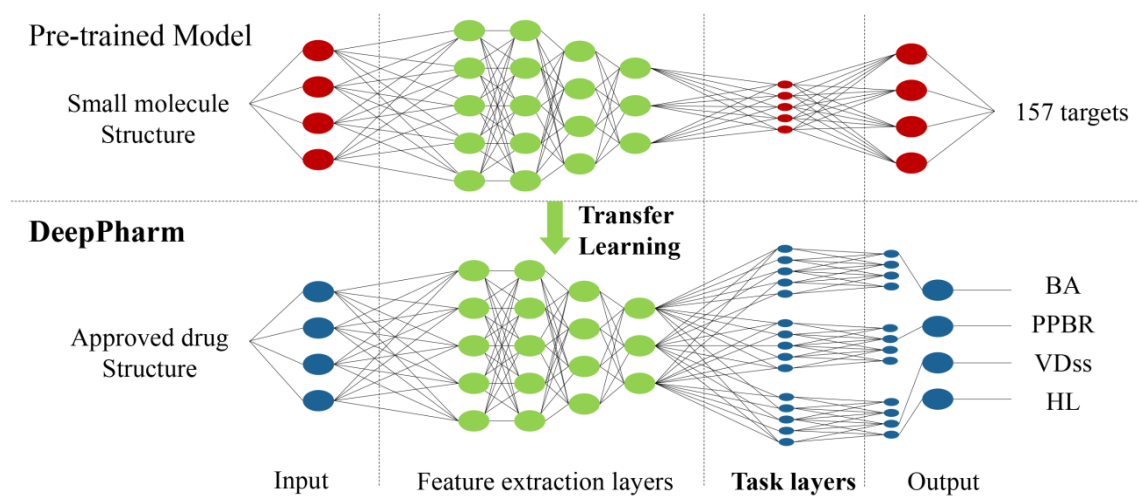

# ABSTRACT:

**Background:** Pharmacokinetic evaluation is one of the key processes in drug discovery and development. However, current absorption, distribution, metabolism, excretion prediction models still have limited accuracy.

**Aim:** This study aims to construct an integrated transfer learning and multitask learning approach for developing quantitative structure-activity relationship models to predict four human pharmacokinetic parameters.

**Methods:** A pharmacokinetic dataset included 1104 U.S. FDA approved small molecule drugs. The dataset included four human pharmacokinetic parameter subsets (oral bioavailability, plasma protein binding rate, apparent volume of distribution at steady-state and elimination half-life). The pre-trained model was trained on over 30 million bioactivity data. An integrated transfer learning and multitask learning approach was established to enhance the model generalization.

**Results:** The pharmacokinetic dataset was split into three parts (60:20:20) for training, validation and test by the improved Maximum Dissimilarity algorithm with the representative initial set selection algorithm and the weighted distance function. The multitask learning techniques enhanced the model predictive ability. The integrated transfer learning and multitask learning model demonstrated the best accuracies, because deep neural networks have the general feature extraction ability, transfer learning and multitask learning improved the model generalization.

**Conclusions:** The integrated transfer learning and multitask learning approach with the improved dataset splitting algorithm was firstly introduced to predict the pharmacokinetic parameters. This method can be further employed in drug discovery and development.

# INTRODUCTION

Early pharmacokinetic parameter prediction is one of the key steps in pharmacokinetic evaluation.[1, 2] In the 1990s, poor ADMET (absorption, distribution, metabolism, excretion and toxicity) properties were the major causes responsible for the attrition of drug candidates in clinical trials.[3] In recent 20 years, advances in preclinical virtual screening methods have been made to predict the pharmacokinetic properties prior to the experiments.[4, 5] These methods included physiologically based pharmacokinetic prediction tools and QSAR (quantitative structure-activity relationship) approaches. In practice, oral bioavailability (BA), plasma protein binding rate (PPBR), apparent volume of distribution at steady-state (VDss) and elimination half-life (HL) were four important and commonly measured human pharmacokinetic parameters. They respectively reflected the fraction reaching systemic circulation of the dose administered, the fraction binding to plasma proteins of the drug in the blood, the *in vivo* distribution of the drug and the time for eliminating half of the initial drug concentration. Currently, the measurements of the four pharmacokinetic parameters still highly rely on the labor-intensive and costly *in vivo* and *in vitro* experiments, which is a heavy burden for the pharmaceutical industry. There has been an increasing interest in the in silico QSAR approaches to optimize and replace the experiments of BA, PPBR, VDss and HL.[6-8]

In recent 20 years, there have been several attempts in QSAR models for the pharmacokinetic parameter prediction. Besides rule-of-thumb which contained simple-rule based classification approaches (e.g. the Lipinski's rule-of-five)[9] and some empirical equations, machine learning techniques were widely adopted in ADME evaluation.[10-19] Machine learning algorithms were used to develop mathematical models by using the existing experimental data. Table 1 summarized published studies which applied a various of machine learning approaches to BA, PPBR, VDss and HL prediction in recent 20 years. Partial least squares regression (PLSR) and support vector machine (SVM) were two popular linear and nonlinear machine learning algorithms in pharmacokinetic parameter prediction. For example, in the year 2011, Hou and co-workers developed the multiple linear regression (MLR) and genetic function approximation (GFA) models to predict the BA in humans. The models were trained on the dataset that included around one thousand drug and drug-like compounds. GFA was the methods that combined the genetic algorithm and the multivariate adaptive regression splines algorithm.[20] In the year 2008, Ma et al. adopted SVM methods with the genetic algorithm and the conjugate gradient methods to identify the PPBR as two categories of positive (≥75%) or negative.[21] In the year 2016, Lombardo and co-workers used PLSR and random forest (RF) algorithms to construct the models for VDss prediction.[22] For HL prediction, in the year 2012, Heidi Kidron and co-workers developed the MLR models on a dataset of 47 compounds.[23] The conventional machine learning methods required considerable and reliable expertise to design the features. But the expertise in molecular structure and pharmacokinetics tend to be insufficient and subjective. On the other hand, lots of molecular features and complex ADME mechanism also need automatic feature extractors. Therefore, there is still room for improvement in QSAR approaches for pharmacokinetic parameter prediction.

Table 1 Recent progresses in BA, VDss, PPBR and HL prediction

| Objective | Data | Machine learning method | Reference |
| --- | --- | --- | --- |
| PPBR | 239 chemicals | Partial least squares | [24] |
| PPBR, VDss | 320 chemicals / 328 chemicals | Multiple linear regression | [25] |
| PPBR | 115 Beta-lactams | Multiple linear regression | [26] |
| PPBR | 1008 chemicals | Multiple linear regression / Artificial neural networks (ANN) / k-nearest neighbors / Support vector machine | [27] |
| PPBR | 853 chemicals | Support vector machine | [21] |
| PPBR | 1242 chemicals | k-nearest Neighbors / Random forest / Support vector machine | [28] |
| PPBR | 1830 chemicals | Random forest / Support vector machine / Cubist | [29] |

| | | Gaussian process | |
| | | Boosting | |
| | | Random Forest | |
| | | Boost Tree | |
| PPBR | 1837 drugs and chemicals | Multiple Linear Regression | 30 |
| | | k-nearest-neighbor | |
| | | Support vector regression | |
| | | Multi-layer Neural Network | |
| VDss | 584 chemicals | Linear squared error model | 31 |
| VDss | 121 drugs | Multiple linear regression | 32 |
| | | Artificial neural network | |
| | | Support vector machines | |
| VDss | 604 chemicals | Decision tree-based regression | 33 |
| VDss | 384 drugs | Mixture discriminant analysis-random forest model | 34 |
| VDss | 698 drugs and chemicals | Partial least-squares | 35 |
| | | Random forest | |
| VDss | 1096 chemicals | Partial Least Squares | 22 |
| | | Random Forest | |
| Vitreal HL | 68 chemicals | Multiple linear regression | 36 |
| Vitreal HL | 47 chemicals | Multiple linear regression | 23 |
| HL | 1105 chemicals | gradient boosting machine | 37 |
| | | support vector regressions | |
| | | local lazy regression and so on | |
| BA | 591 chemicals | Stepwise regression | 38 |
| BA | 232 drugs | Adaptive least squares | 39 |
| BA | 167 drugs | Artificial neural networks | 40 |
| BA | 302 drugs | Partial least squares | 41 |
| BA | 768 chemicals | Rule-based approaches | 42 |
| BA | 772 chemicals | Multiple linear regression | 43 |
| BA | 766 chemicals | Support vector machine | 21 |
| BA | 1014 chemicals | Multiple linear regression<br>Genetic function approximation | 20 |
| BA | 805 drugs and chemicals | Multiple linear regression<br>Partial least squares regression<br>Support vector machine | 44 |

Recently, deep learning, which is usually accepted as deep neural networks (DNN), has been widely used to make advances in many fields of academia and industry.[45-48] The key aspect of deep learning is the general-purpose feature extraction procedure, which can automatically transform raw data into higher-level features. The intricate structure and minute variations of the high-dimensional data can be discovered by deep learning without human expertise.[49, 50] In recent years, many successes have also been made by deep learning in drug discovery and formulation development.[51, 52] For examples, in the year 2013, compared with other machine learning methods, deep learning yielded good performance in aqueous solubility prediction based on the undirected graphs.[53] In the year 2015, it was found that deep learning could reach or exceed the performance of the RF algorithms on a series of Merck's drug discovery datasets.[54]

The most difficulty in ADME evaluation is the lack of sufficient and high quality data. By utilizing the massive bioactivity data to find out the common low level features, the accuracies of ADME models can be enhanced by transfer learning. Transfer learning aims to transfer known knowledge to the target domain to improve the learning.[55] Leveraging the common features learned from the similar source domain, transfer learning has been demonstrated to be able to develop the models without learning from scratch.[56-58] Recently, in the medical field, transfer learning has made great progress in identifying optical coherence tomography images to aid medical decision making.[59] Activity-related chemical characteristics of molecules can be discovered by deep learning on the massive molecular structure and bioactivity measurements on the basis of the assumption that similar molecules have similar properties.[60] The chemical characteristics can be transferred to the pharmacokinetic models by transfer learning. Furthermore, by uncovering the low level common features among multiple tasks simultaneously, it was found that multitask learning was able to improve the model generalization.[61] Multitask learning has been reported to be applied to drug discovery.[62-65]

Nowadays, there are over 30 million bioactivity data which include hundreds of thousands of molecules. These bioactivity data include the molecular structure and the bioactivity measurements. Bioactivity measurements described the activities of compounds against the biological targets. Rather than developing the ADME model from a model with initialization weights, the model could be obtained through retraining the weights which have been pre-optimized on a large bioactivity database for the pharmacokinetic parameter prediction (Figure 1).

In this paper, transfer learning and multitask learning methods were applied to the pharmacokinetic parameter prediction. In detail, a pharmacokinetic dataset comprising four critical human pharmacokinetic parameters was built and a large bioactivity dataset containing over 30 million entries was introduced. The specific evaluation criterion suitable for pharmacokinetics and the automatic dataset splitting algorithm were developed. The deep neural networks were trained by using the integrated transfer learning and multitask learning approach. Compared with other machine learning methods, higher accuracies and stronger generalization of the present models were shown by the results.

## METHODS

**1. Pharmaceutical Data**

*1.1. Pharmacokinetic Dataset*

The pharmacokinetic dataset contained 1104 approved drugs, which was a union of the four subsets. The subsets included a BA subset of 410 molecules, a PPBR subset of 769 molecules, a VDss subset of 412 molecules and a HL subset of 969 molecules. The approved drug list was obtained from U.S. Food and Drug

Administration. The pharmacokinetic information of the drugs was retrieved from Drugbank database.[66] The information was manually normalized to the quantitative data. In this study, the data determined in healthy adults were adopted rather than the elder, children and patients. The data of instant release dosage forms were reserved rather than sustained release dosage forms.

*1.2. Bioactivity Dataset for Developing the Pre-trained Model*

The pre-trained model for transfer learning was trained on the large bioactivity dataset. This large bioactivity dataset contained more than 30.9 million entries of qualitative data and was obtained from Deepchem.[67] This dataset included 486904 molecules and 157 targets. Three subsets of PCBA, MUV and Tox21 in this dataset were collected from PubChem database and the 2014 Tox21 Data Challenge by DeepChem.[68]

## 2. Representation of molecular structure

The molecules used in this study were characterized by Extended-Connectivity Fingerprints (ECFPs).[69] ECFPs were designed for developing QSAR models. ECFPs have been widely used in the fields of cheminformatics and bioinformatics. In this study, the ECFPs were generated by using the open source package RDKit.[70] The Canonical SMILES of the molecules were used as the input of RDKit, the length and radius of the ECFPs were set to 1024 and 2. The Canonical SMILES were obtained from PubChem database.[68]

Because ECFPs are binary data which are not suitable for the molecular similarity calculating for dataset splitting, the eight molecular descriptors commonly used were adopted to calculate the similarity for dataset splitting. The eight molecular descriptors of the approved drugs were obtained from PubChem database.[68] The descriptors included Molecular Weight, Topological Polar Surface Area, Rotatable Bond Count, Hydrogen Bond Donor Count, Hydrogen Bond Acceptor Count, Heavy Atom Count, Complexity and Covalently-Bonded Unit Count.

## 3. Three-dataset Splitting Strategy

The three-dataset splitting strategy was widely accepted in the field of machine learning. In this study, the pharmacokinetic dataset was split into three subsets (training/validation/test subsets). The training subset was for training models. The validation subset was for tuning the hyper-parameters to find the best model. The test subset was for testing the model performances on an external dataset. The training, validation and test subsets contained 664, 220 and 220 drugs, respectively.

# 4. Metric Criteria

## 4.1. Pharmacokinetic Task

In machine learning, correlation coefficient and determinant coefficient were two common metric criteria used to evaluate the linear relationship between the real values and the predicted values. However, both of them can't well reflect the real performances of the QSAR models in ADME evaluation. The absolute error (AE) was introduced to evaluate the accuracies of the machine learning models. The AE was defined as the difference between a real value and a predicted value. In pharmacokinetic parameter prediction, if the AE is no more than 10%, it can be thought of a successful prediction. Therefore, the accuracy is the percentage of the successful predictions in all predictions:

$$Accuracy = \frac{Number(|AE| \leq 0.1)}{AllPredictions}$$

In addition, the MAE (mean absolute error) was also used as metrics and defined as follow:

$$MAE = \frac{\sum_{i=1}^{n}|y_i^{pred} - y_i^{label}|}{n}$$

Where, n is the number of samples, $y^{pred}$ are the prediction values, $y^{label}$ are the experimental values.

## 4.2. Bioactivity Task of the Pre-trained Model

In machine learning, receiver operating characteristic (ROC) curve was commonly used to evaluate performances for classification problems. Though the bioactivity dataset contains 30.9 million entries, only 0.433 million entries (about 1.4% of the total entries) are active and the residual entries are negative. Thus, a recall value was more suitable than ROC curve in this case. The recall was defined as:

$$Recall = \frac{TP}{TP + FN}$$

Where, TP is true positive, FN is false negative.

**5. Hyper-parameters of Conventional Machine Learning Algorithms**

Regression models were constructed on the basis of the five machine learning algorithms containing PLSR, SVM, ANN, RF and k-nearest neighbors (k-NN). These QSAR models were built using the sci-kit learn package in Python programming language.[71] For BA, PPBR, VDss and HL, in PLSR, the number of components was chosen as 5, 6, 2 and 5. In ANNs, the networks containing 1 hidden layer with 200, 600, 500 and 400 hidden nodes were adopted. In RF, the maximum depths of the tree of 10, 15, 10, 5 were used. In k-NN, 5, 2, 20, 20 were set as the numbers of neighbors.

**6. An Integrated Transfer Learning and Multitask Learning Approach**

*6.1. Multitask Learning Model*

The multitask learning feed-forward neural network was developed on the pharmacokinetic dataset. The model was trained on the basis of the DeepLearning4j framework in Java programming language. This network contained 10 dense layers. The hidden nodes in the 10 layers were set from 1000 to 100 with a drop of 100 nodes per layer. The epoch was set to 100. The learning rate was set to 0.1. The $\beta_1$, $\beta_2$ and $\lambda$ were set to 0.5, 0.999 and 0.01, respectively. The tanh was chosen as the activation function except the last layer with the sigmoid activation function.

*6.2. Pre-trained Model*

The pre-trained model was developed on the bioactivity dataset. This neural network was trained on the DeepLearning4j framework in Java programming language. 11 dense layers including 10 feature extraction layers and 1 task layer were adopted in this network. The hidden nodes in the feature extraction layers were set from 1000 to 100 with a drop of 100 nodes per layer. The task layer contained 1000 hidden nodes. The epoch was set to 5. The learning rate was set to 0.01. The $\beta_1$, $\beta_2$ and $\lambda$ were set to 0.5, 0.999 and 0.01, respectively. The tanh was chosen as the activation function except the last layer with the sigmoid activation function.

*6.3. DeepPharm Models*

Three feed-forward neural networks (DeepPharm-BA, DeepPharm-PPBR and DeepPharm-VDss&HL models) were developed. The models were trained using the integrated transfer learning and multitask learning approach (DeepPharm) on the DeepLearning4j framework in Java programming language. The framework of this approach was illustrated in Figure 1. The three deep neural networks contained 11 dense layers of 10 feature extraction layers and 1 task layer. The hidden nodes in the feature extraction layers were set from 1000 to 100 hidden nodes with a drop of 100 nodes per layer. The internal weights of the feature extraction layers of the three networks were transferred from the pre-trained model and retrained on the pharmacokinetic dataset. 96 epoch and a task layer of 1000 hidden nodes were used to develop the DeepPharm-BA model. 52 epoch and a task layer of 1000 hidden nodes were used to develop the DeepPharm-PPBR model. 96 epoch and a task layer of 100 hidden nodes were used to develop the DeepPharm-VDss&HL model. The learning rate was set to 0.03, the $\beta_1$, $\beta_2$ and $\lambda$ were set to 0.5, 0.999 and 0.01. The tanh was chosen as the activation function except the last layers with the sigmoid activation function.

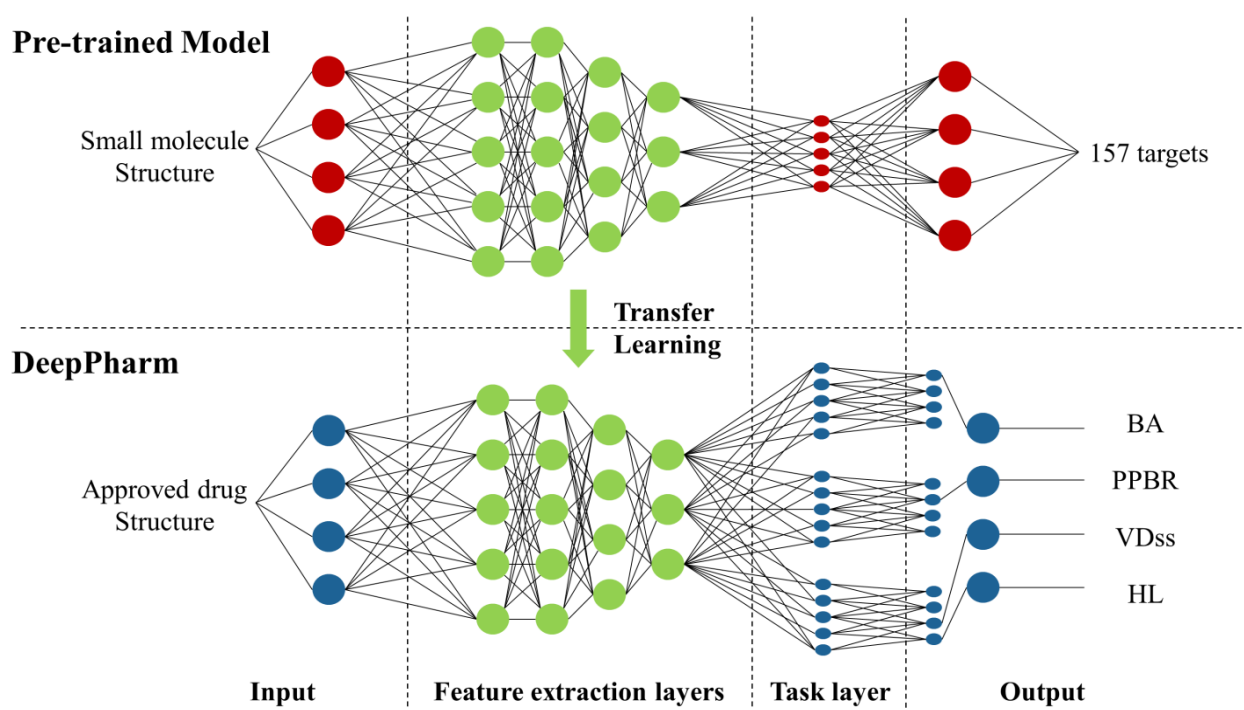

Figure 1 The framework of the integrated transfer learning and multitask learning approach (DeepPharm)

## RESULTS

### 1. Dataset Splitting Algorithms

The data ranges and distributions of the four pharmacokinetic parameters were different. The ranges of BA and PPBR were from 0% to 100%. The VDss and HL were no more than 2000L and 168h. The distributions of VDss and HL data were extremely imbalance. Over 75% VDss data concentrated on the range of 0-100L. Over 60% HL data were less than 10 hours. Therefore, an investigation of the automatic dataset splitting algorithms for the pharmacokinetic dataset splitting was required.

Firstly, the random dataset splitting was carried out. An analysis of the performance of dataset splitting algorithms was implemented. The split training, validation and test subsets were equally divided into 10 groups according to the subset ranges, respectively. Subset error (SE) was introduced to evaluate the representativeness of the split subsets. SE was defined as:

$$SE = \frac{\sum_{i=1}^{n}(MaxFreq_i - MinFreq_i)}{n}$$

where, $MaxFreq$ is the maximum frequency of the data in each group among the training, validation and test subsets, $MinFreq$ is the minimum frequency of the data in each group among the training, validation and test subsets. $n$ is the number of the groups and set to 10. The results of the large SE values indicated that random splitting methods couldn't select the representative subsets from the pharmacokinetic dataset.

Recently, an automatic splitting algorithm (the MD-FIS algorithm) selecting the representative subsets from the formulation dataset was published by us.[51, 52] But the pharmacokinetic data were different from the formulation data, the pharmacokinetic data were imbalanced in the distributions of the eight molecular descriptors and the pharmacokinetic parameters. Therefore, a new MD-FIS algorithm (MD-FIS-WD) with the Weighted Distance function was developed in R programming language for improving the splitting performance. The weighted distance function was defined as:

$$Distance = w_1 d + w_2 p$$

Where, $w_1$ and $w_2$ can control the proportion of the molecular descriptors and the pharmacokinetic parameters, $d$ is the molecular descriptor and $p$ is the pharmacokinetic parameter. Finally, $w_1$ and $w_2$ were set to 0.7 and 0.3 to get the best splitting performance. The SE values of BA, PPBR, VDss and HL were 4.21, 3.18, 4.32 and 3.87, respectively. The SE values yielded by MD-FIS-WD were smaller than those got by random splitting methods and the MD-FIS algorithm considering only the molecular descriptors. The training, validation and test subsets split by the MD-FIS-WD algorithm were used to develop machine learning models.

## 2. Conventional Machine Learning Methods

Five machine learning algorithms including PLSR, SVM, ANNs, RF and k-NN were introduced to develop QSAR models for pharmacokinetic evaluation. Table 2 showed the accuracies of the five machine learning models on training, validation and test sets. On the test set, for BA, the accuracies of the five machine learning models were around 20% with the lowest accuracy of 15.07% of ANN and the highest accuracy of 23.29% of SVM. The RF model got an accuracy of 21.92% inferior to the SVM model. For PPBR, the accuracies of the five models were around 30%. The ANN model got the best results of 33.56%. The SVM, RF and k-NN models got the same accuracies of 31.54% closed to the ANN model. For VDss, the five model accuracies were from 40.66% to 60.44%. The k-NN model got the highest accuracy of 60.44% among them. The SVM model got the second highest accuracy of 52.75%. For HL, the lowest accuracy was 52.00% of the ANN model and the highest accuracy was 66.29% of the RF model. The k-NN model got an accuracy of 65.71% closed to the RF model.

Table 2 Accuracy and MAE values of BA, PPBR, VDss and HL on training, validation and test sets

| Characteristic | Machine Learning | Training set | | Validation set | | Test set | |
|---|---|---|---|---|---|---|---|
| | | Accuracy | MAE | Accuracy | MAE | Accuracy | MAE |
| BA | PLSR | 95.45 | 0.0341 | 35.04 | 0.2476 | 20.55 | 0.3144 |
| | SVM | 90.45 | 0.0435 | 29.91 | 0.2707 | **23.29** | 0.3418 |
| | ANN | 96.82 | 0.0217 | 30.77 | 0.2736 | 15.07 | 0.3749 |
| | RF | 43.64 | 0.1363 | 27.35 | 0.2459 | 21.92 | 0.2756 |
| | k-NN | 29.09 | 0.1990 | 23.93 | 0.2543 | 19.18 | 0.3266 |
| | Multi-task | 92.27 | 0.0318 | 37.61 | 0.2518 | 25.00 | 0.3019 |
| | DeepPharm | 86.36 | 0.0503 | 32.48 | 0.2570 | **27.78** | 0.3122 |
| PPBR | PLSR | 92.98 | 0.0416 | 31.10 | 0.2337 | 30.87 | 0.2759 |
| | SVM | 98.90 | 0.0213 | 29.27 | 0.2633 | 31.54 | 0.2883 |
| | ANN | 99.56 | 0.0130 | 34.15 | 0.2509 | **33.56** | 0.2789 |
| | RF | 68.64 | 0.0882 | 27.44 | 0.2325 | 31.54 | 0.2368 |
| | k-NN | 69.08 | 0.0951 | 43.90 | 0.2226 | 31.54 | 0.2604 |
| | Multi-task | 80.09 | 0.0703 | 46.58 | 0.1897 | 42.86 | 0.2196 |
| | DeepPharm | 41.37 | 0.2049 | 41.61 | 0.2528 | **44.22** | 0.2481 |
| VDss | PLSR | 97.36 | 0.0327 | 68.09 | 0.1187 | 49.45 | 0.1759 |
| | SVM | 99.56 | 0.0241 | 67.02 | 0.1213 | 52.75 | 0.1766 |
| | ANN | 98.68 | 0.0210 | 53.19 | 0.1501 | 40.66 | 0.2020 |
| | RF | 83.26 | 0.0587 | 52.13 | 0.1387 | 48.35 | 0.2092 |
| | k-NN | 81.94 | 0.0791 | 77.66 | 0.1124 | **60.44** | 0.1917 |
| | Multi-task | 98.24 | 0.0119 | 72.34 | 0.1075 | 61.11 | 0.1732 |
| | DeepPharm | 96.04 | 0.0270 | 78.72 | 0.0982 | **63.33** | 0.1735 |
| HL | PLSR | 98.23 | 0.0269 | 68.97 | 0.1114 | 56.00 | 0.1464 |
| | SVM | 95.81 | 0.0363 | 68.39 | 0.1085 | 56.00 | 0.1476 |
| | ANN | 96.77 | 0.0215 | 51.72 | 0.1338 | 52.00 | 0.1514 |
| | RF | 88.06 | 0.0486 | 78.16 | 0.0925 | **66.29** | 0.1269 |
| | k-NN | 87.74 | 0.0508 | 81.61 | 0.0879 | 65.71 | 0.1338 |
| | Multi-task | 89.19 | 0.0520 | 77.59 | 0.0926 | **68.39** | 0.1259 |
| | DeepPharm | 94.19 | 0.0327 | 78.74 | 0.0873 | 66.67 | 0.1216 |

## 3. The Integrated Transfer Learning and Multitask Learning Approach

### 3.1. Multitask Learning Model

Multitask learning techniques were implemented to develop a deep neural network. Conventional multitask learning methods cannot solve the problem that the training set is imbalanced with missing label values. A dynamic weighted cost function was introduced to give different weights for different pharmacokinetic parameters.

$$\text{cost} = w_1 c_1 + w_2 c_2 + w_3 c_3 + w_4 c_4$$

Where, $w_1$, $w_2$, $w_3$ and $w_4$ can control the costs of BA, PPBR, VDss and HL, respectively. Finally, the weights were set to $w_1 : w_2 : w_3 : w_4 = 3 : 1 : 9 : 1$. The results of the multitask learning deep neural network were shown in table 2. For all pharmacokinetic parameters, on the test set, the accuracies of the multitask learning model were higher than the five conventional machine learning algorithms.

### 3.2. Pre-trained Model

The integrated transfer learning and multitask learning approach was implemented for better model generalization and predictive ability. A pre-trained model was developed on the large bioactivity dataset. The bioactivity dataset was randomly divided into two parts of the training set and the validation set. The results showed the recall values were 49.51% on the training set, and 32.02% on the validation set. Furthermore, a weighted cost function was introduced to enhance the proportion of the costs of the active data.

$$\text{wcost} = \begin{cases} \dfrac{w_1 (pv - lv)^2}{2} & \text{if } label = 1 \\ \dfrac{w_2 (pv - lv)^2}{2} & \text{if } label = 0 \end{cases}$$

Where, $w_1$ and $w_2$ can control the costs of active data ($label = 1$) and negative data ($label = 0$), $pv$ is the predictive value, $lv$ is the label value. Finally, $w_1 : w_2 = 100 : 1$ was used, which resulted in the recall values of 87.42% on the training set, 85.05% on the test set.

### 3.3. DeepPharm Models

The three deep neural networks (DeepPharm-BA, DeepPharm-PPBR and DeepPharm-VDss&HL) were developed using the integrated transfer learning and multitask learning approach. Usually, a consensus model can take the advantages of various models [20, 72]. The consensus model (DeepPharm model) took the optimal prediction for each of the four pharmacokinetic parameters from the three networks. The accuracies of the DeepPharm model were shown in Table 2. On the test set, the DeepPharm model got the highest accuracies of 27.78%, 44.22% and 63.33% in BA, PPBR and VDss prediction. For HL, on the test set, the DeepPharm model got the closed result of 66.67% to the multitask model result of 68.39% and was superior to other conventional machine learning methods. Table 3 showed the DeepPharm model performance on the basis of the metric criteria using different benchmarks of 10%, 20% and 30%. When the benchmark was set to 20%, the accuracies of the DeepPharm model were more than 75% in VDss and HL prediction.

**Table 3** Accuracies across different metric criteria using the benchmarks of 10%, 20% and 30% of the DeepPharm model for BA, PPBR, VDss and HL on the test subsets

| Characteristic | % ≤ 10% error | % ≤ 20% error | % ≤ 30% error |
| --- | --- | --- | --- |
| BA | 27.78 | 41.67 | 51.39 |
| PPBR | 44.22 | 59.86 | 67.35 |
| VDss | 63.33 | 75.56 | 78.89 |
| HL | 66.67 | 81.03 | 90.23 |

## DISCUSSION

The lack of sufficient and high quality data is one factor of the difficulties in ADME evaluation. The high cost of clinical trials results in the lack of reliable pharmacokinetic data. This problem has been mentioned in many previous studies.[22, 44, 54, 73, 74] Generally, the validation and the test sets should have the same probability distribution with the training set. Random splitting may lead to an imbalanced dataset splitting on a small dataset. Many studies used the k-fold or leave-one-out (LOO) cross-validation methods to evaluate the model performance, while the k-fold and LOO cross-validation methods may get overoptimistic results.[75] In this study, the MD-FIS-WD automatic dataset splitting algorithm was developed to handle the pharmacokinetic dataset splitting problem. Considering both molecular structure and pharmacokinetic parameter distributions, this algorithm works well according to the data splitting results. Moreover, molecular representation of compounds is also an important factor to predict the ADME parameters. In our manuscript, the ECFP_4 molecular fingerprint were used as the input of both bioactivity dataset and PK predictions.

The third factor is the algorithm. As shown in Table 1, in recent 20 years, several QSAR models have been trained for the pharmacokinetic parameter prediction. They performed better than the rule-of-thumb, such as simple rule-based classification methods and empirical equations. These models were learned from *in vitro* and *in vivo* experimental data by a serious of machine learning approaches. The physicochemical properties, molecular descriptors and molecular fingerprints were used for the representation of molecular structure.[76] For example, in the year 2012, Xu et al. compared multiple linear regression to partial least squares regression and support vector machine for BA prediction. The models were generated using molecular descriptors.[44] In the year 2017, Cao and co-workers presented models developed by RF, SVM, Cubist, Gaussian process, and Boosting for PPBR prediction. The molecular descriptors used in the work were selected by non-dominated sorting genetic algorithms and partial least squares regression.[29] In the year 2015, Alex et al. applied decision tree methods to VDss prediction based on molecular descriptors and tissue: plasma partition coefficients.[33] In the year 2016, Lu et al. developed QSAR models based on a dataset of 1105 organic chemicals for HL prediction.[37] However, conventional machine learning algorithms highly relied on feature engineering. Feature engineering was time-consuming and challenging due to human experts' subjective and vague experiences, which resulted in bad model performance. In our study, deep learning can automatically extract the critical features or molecular descriptors from the raw data without the need of feature engineering. The key aspect of deep learning is the general-purpose feature extraction procedure, which can automatically transform raw data into higher-level features.

Moreover, four independent variables in PK-related property (BA, PPBR, VDss and HL) were predicted by integrated transfer learning and multi-tasks approach. Applying multitask learning, the deep neural network performed better than other single-task machine learning methods. Because multitask learning could utilize the data of multiple tasks to discover the correlations and distinguish the differences among multiple tasks. Leveraging the bioactivity data, transfer learning further enhanced the model generalization ability. Transfer learning achieved good prediction ability by reusing the features of the pre-trained model as the initial model weights. Furthermore, transfer learning techniques could save computing resources and time for model optimization as well. The good model generalization was showed by the external unseen test dataset. With the increase of reliable pharmacokinetic data, future models may benefit more by using this approach.

In this research, we found that the nonlinear machine learning models performed better than the linear PLSR models. Obviously, the relationships between the molecular structure and the pharmacokinetic parameters are non-linear. In many cases, the random forest models are superior to other conventional machine learning algorithms. Random forest has more complex non-linear model structure than other methods by aggregating decision trees, which could construct a more complex function mapping. Random forest is more likely to depict the nonlinear relationship between the molecular structure and the pharmacokinetic parameters. In addition, generally, deep neural networks directly optimized on a large dataset would get higher accuracy than transfer learning because the models could be directly trained for the specific task. However, the main difficulty in ADME evaluation was the small dataset with imbalanced input space. Compared to the models developed from the weights using random initialization, transfer learning could obtain more accurate models on small training datasets. Furthermore, the feature extraction layers of a pre-trained model are very important in transfer learning. A pre-trained model trained on large and high quality relevant dataset would enhance the performance of the target model.

As described above, we demonstrated the integrated transfer learning and multitask learning approach for BA, PPBR, VDss and HL prediction, which provided a reference for the ADME evaluation and may accelerate the drug discovery and development. BA, as a complex pharmacokinetic parameter determining the administered dose, depends highly on several absorption, transport and metabolism mechanisms in gastrointestinal absorption and excretion process, in gut wall first-pass process and in hepatic first-pass process. PPBR is an important pharmacokinetic parameter. Generally, it is assumed that only free drugs could pass through the vascular wall. The combined drug in the blood is a reservoir, which would cause drug overdose and toxicity. PPBR has an impact on the VDss. VDss is a theoretical parameter which reflects *in vivo* distribution of a drug. A low VDss value generally indicates that the drug concentrate in the vascular and a high VDss value indicates extensive drug distribution in the body. HL is influenced by clearance and VDss. HL describes the time for decreasing half of the initial plasma drug concentration. HL is an important pharmacokinetic parameter. HL determines the administered frequency of a drug. Considering the importance of preclinical ADME evaluation, more applications of this approach may be carried out for the prediction of other pharmacokinetic parameters. Besides pharmacokinetic parameter prediction, more attempts of the integrated transfer learning and multitask learning approach may exceed the field of pharmacokinetics in the future.

## CONCLUSIONS

In this paper, an integrated transfer learning and multitask learning approach with the MD-FIS-WD dataset splitting algorithm was successfully established to develop QSAR models to predict four critical human pharmacokinetic parameters on the approved drug dataset. The final DeepPharm model showed the best performance and generalization ability than other conventional machine learning approaches because deep neural networks have the general feature extraction ability and transfer learning improves the model generalization. Our integrated transfer learning and multitask learning approach may reveal a broad prospect of the application in future pharmaceutical researches.

## ACKNOWLEDGMENTS

Current research is financially supported by the University of Macau Research Grant (MYRG2016-00038-ICMS-QRCM, MYRG2016-00040-ICMS-QRCM and MYRG2017-00141-FST), Macau Science and Technology Development Fund (FDCT) (Grant No. 103/2015/A3), 2017 Sub-project 6 of National Major Scientific and Technological Special Project for "Significant New Drugs Development" of the Ministry of Science and Technology of China (2017ZX09101001006) and the National Natural Science Foundation of China (Grant No. 61562011).